\documentclass[runningheads]{llncs}
\usepackage{graphicx}
\usepackage{caption}
\usepackage{subcaption}
\usepackage{hyperref}
\usepackage{xcolor}
\usepackage{url}
\usepackage{arydshln} 
\usepackage{amssymb} 
\usepackage{booktabs}
\usepackage{float}
\usepackage{footnote}
\usepackage{multirow}
\usepackage{tikz}
\usepackage{amsmath}
\usepackage{color}
\usepackage{pgfplots}
\usepackage{pgf-umlsd}
\usepackage{ifthen}
\usepackage{forest}
\usepackage[misc]{ifsym}
\makeatletter
\renewcommand\section{\@startsection{section}{1}{\z@}%
                       {-8\p@ \@plus -4\p@ \@minus -4\p@}%
                       {6\p@ \@plus 4\p@ \@minus 4\p@}%
                       {\normalfont\large\bfseries\boldmath
                        \rightskip=\z@ \@plus 8em\pretolerance=10000 }}
\renewcommand\subsection{\@startsection{subsection}{2}{\z@}%
                       {-8\p@ \@plus -4\p@ \@minus -4\p@}%
                       {6\p@ \@plus 4\p@ \@minus 4\p@}%
                       {\normalfont\normalsize\bfseries\boldmath
                        \rightskip=\z@ \@plus 8em\pretolerance=10000 }}
\renewcommand\subsubsection{\@startsection{subsubsection}{3}{\z@}%
                       {-4\p@ \@plus -4\p@ \@minus -4\p@}%
                       {-1.5em \@plus -0.22em \@minus -0.1em}%
                       {\normalfont\normalsize\bfseries\boldmath}}
\makeatother

\begin{document}
\title{Auto-Classifier: A Robust Defect Detector Based on an AutoML Head\thanks{This work was supported by NOVA LINCS (UIDB/04516/2020) with the financial support of FCT-Fundação para a Ciência e a Tecnologia, through national funds, and partially supported by project 026653 (POCI-01-0247-FEDER-026653) INDTECH 4.0 – New technologies for smart manufacturing, cofinanced by the Portugal 2020 Program (PT 2020), Compete 2020 Program and the European Union through the European Regional Development Fund (ERDF).}}
\titlerunning{Auto-Classifier: A Robust Defect Detector Based on an AutoML Head}
\toctitle{Auto-Classifier: A Robust Defect Detector Based on an AutoML Head}

\author{Vasco Lopes {\Letter} \and
Lu\'is A. Alexandre}

\authorrunning{V. Lopes and L. A. Alexandre}

\tocauthor{Vasco~Lopes (NOVA~LINCS,~Universidade~da Beira~Interior),
Luís~A.~Alexandre (NOVA~LINCS,~Universidade~da Beira~Interior),
}

\institute{NOVA LINCS, 
Universidade da Beira Interior\\
Rua Marqu\^es d' \'Avila e Bolama, 6201-001 Covilh\~a, Portugal\\
\email{$\{$vasco.lopes,luis.alexandre$\}$@ubi.pt}}

\maketitle   
\setcounter{footnote}{0}

\begin{abstract}
The dominant approach for surface defect detection is the use of hand-crafted feature-based methods. However, this falls short when conditions vary that affect extracted images. So, in this paper, we sought to determine how well several state-of-the-art Convolutional Neural Networks perform in the task of surface defect detection. Moreover, we propose two methods: CNN-Fusion, that fuses the prediction of all the networks into a final one, and Auto-Classifier, which is a novel proposal that improves a Convolutional Neural Network by modifying its classification component using AutoML. We carried out experiments to evaluate the proposed methods in the task of surface defect detection using different datasets from DAGM2007. We show that the use of Convolutional Neural Networks achieves better results than traditional methods, and also, that Auto-Classifier out-performs all other methods, by achieving 100\% accuracy and 100\% AUC results throughout all the datasets.

\end{abstract}

\section{Introduction}
Visual inspection of products is crucial to ensure customer requirements and a longer product life, by removing imperfections or defects that can lead to problems like rust, sharp edges or visually deficient products. Industrial quality control and visual inspection require extreme attention to detail and are usually performed by humans, meaning that it is prone to error, requires training, and its a time-consuming task that needs to be repeated countless times in modern factories \cite{mital1998comparison}. Systems capable of providing a way to automate such processes, either by completely removing the need for human labour or by complementing the work conducted by humans, are essential to reduce costs and improve product quality \cite{malamas2003survey}. Thus, the goal of inspection systems is to rapidly and precisely detect, classify or segment defective areas in images. However, such systems are scarce due to difficulties in acquiring real data to train Artificial Intelligence (AI) algorithms, and because industry floors require continuous changes, meaning that having controlled and unchanged environments is challenging. Traditional methods often rely on extracting hand-craft features from images, in order to represent defects and anomalies \cite{scholz2012automated,4631331} and can be categorized into: statistical, structural or filter based. These methods are capable of detecting and or segmenting defects in images if they are acquired in a controlled environment \cite{xie2008review}. However, these are not capable of solving the same kind of problems when applied to images with complex textures, or if the acquire data suffers slight changes or contains noise.

To mitigate the aforementioned problems, some approaches evaluate the use of deep learning to tackle the problem of detecting defects \cite{weimer2016design}, more specifically, Convolutional Neural Networks (CNNs), due to the excellent results that they achieve in a multitude of tasks related to image analysis \cite{lecun2015deep,schmidhuber2015deep}. The use of CNNs is particularly good in the task of defect detection, because they can learn to be robust to the presence of noise and different conditions, such as light and rotation \cite{sohn2012learning,cheng2016learning}, meaning that a robust CNN that can correctly classify or detect defects, and can be invariant to the problems that undermine traditional approaches. 

In this paper, we evaluate how several state-of-the-art CNNs perform in the task of surface defect detection, and propose two different approaches: 1) a CNN-fusion method, that averages the classification of all individual CNNs into a final classification, and 2) a Auto-Classifier detector, which is a novel method that integrates the use of an AutoML head to complement a CNN, by using the feature extractor of the CNN and creating a new classifier upon it. We validate our proposals in the task of surface defect detection using DAGM2007 datasets \cite{wieler2007weakly}, and show that the propose methods can improve upon the state-of-the-art. The code for all proposed methods, as well as the used data set partitions is freely available allowing for free use and fair comparisons\footnote{www.github.com/VascoLopes/AutoClassifier}.

The rest of this paper is organized as follows: the next section discusses the state-of-the-art and related work regarding defect detection and AutoML; section \ref{sec:proposed} presents the proposed methods; section \ref{sec:experiments} contains the experiments and discussion, while the final sections contain the conclusions.

\section{Related Work}


\noindent \textbf{AutoML:} 
The field of AutoML is a domain of expertize whose aim is to develop methods and tools to provide efficient mechanisms that can be used by virtually anyone to design tailor-made machine learning algorithms to their problems \cite{hutter2019automated}. Designing ML algorithms for specific problems can be a difficult task, as it can have many design choices that are both dependent and independent from one another. Designing efficient ML algorithms takes years of expertize and trial and error, as many optimization problems usually rely on the user, which makes it very difficult for non-expert users to do. So, AutoML is an important approach to bring machine learning algorithms closer to non-experts, but also to integrate it with other technologies, to create new and more efficient approaches for several problems. The difference between AutoML and traditional machine learning workflows, is that AutoML intends to remove all the steps between the data acquisition and getting the final model, which usually involve data processing, feature extraction and model selection, from the user.

Over the years, many AutoML methods have been proposed  \cite{kotthoff2017auto,feurer2019auto,mendoza-automlbook18a,H2OAutoML}. However, our proposals are not related to AutoML algorithms, but with the combination of the power that AutoML provides to create optimal, tailor-made algorithms to solve the tasks at hand. Closer to our Auto-Classifier, \cite{kocbek2019automated} uses AutoML to fully design a method to detect railway track defects, while our Auto-Classifier, uses AutoML to complement the feature extraction power of a CNN, by coupling a new classifier with a modified CNN. 

Our work relates with Neural Architecture Search (NAS), which is a subset of AutoML that focus on automatically design deep neural architectures. NAS was initially formulated as a reinforcement learning problem, where a controller is trained over-time to sample more efficient architectures \cite{DBLP:journals/corr/ZophL16}, requiring over 60 years of computation. A follow-up work, improved upon the base work by performing a cell-based search, where cells, which conduct some operations, are replicated to form a complete CNN \cite{Zoph_2018}. In \cite{DBLP:conf/iclr/BakerGNR17}, the authors use Q-learning to train the sampler agent. Using a similar approach, the authors of \cite{DBLP:conf/cvpr/ZhongYWSL18}, perform NAS by sampling blocks of operations instead of cells/architectures, which can then be replicated to form networks. More recently, ENAS \cite{ENAS}, used a controller to discover architectures by searching for an optimal subgraph within a large computational graph, requiring only a few computational days to build a final architecture. DARTS, a gradient-based method, showed that by performing a continuous relaxation of the parameters, they could be optimized using a bi-level gradient optimization \cite{liu2018darts}. DARTS was then improved using regularization mechanisms \cite{zela2020understanding}.

The main difference between NAS and our work is that while NAS focus on designing an entire network, our focus with Auto-Classifier, is to improve upon a CNN that yields good results by leveraging the power of AutoML to design a new classifier component. This search extends the use of CNNs with other types of classifiers, and is faster than designing entire machine learning algorithms from scratch using NAS. This can be seen in the experiments section, where the search for a new classifier was extremely fast, by having AutoML to search for only two hours.

\paragraph{}
\noindent \textbf{Surface Defects:} Different types of surface defects include cracks, which can happen in a panoply of surfaces. In \cite{7926704}, the authors focus on detecting cracks in power plants in a private dataset, using a CNN for semantic segmentation. Similarly, in \cite{li2017recognition}, a bridge cracks detection algorithm is proposed. This method uses active contours and Canny Edge detector to find the defects, and an SVM for classification. In \cite{zhang2016road}, the authors trained a CNN to solve the problem of detecting road cracks, using a dataset of 500 images acquired using a smartphone.

In the task of producing gravure cylinders, it is common to have defects like holes, so, the authors of \cite{villalba2019deep} proposed a method that uses a CNN to classify images acquired by a high-resolution camera. The method achieved an accuracy rate of 98.4\% on a private dataset. Differently, the authors of \cite{garcia2018multi}, proposed a method to conduct surface quality control, by using cutting force, vibration, and acoustic emission signals information of a CNC machine. By decomposing the signals into time series, a predictor was capable of predicting the surface finish. 

Using DAGM2007 set of problems, the author of \cite{weimer2016design} propose to evaluate the performance of 3 CNNs, with different network specifications, which achieve results between 96\% and 99\% accuracy. Traditional methods that rely on extensive feature extraction were also studied. However, their results fall short when compared to deep learning techniques. In \cite{scholz2012automated}, a method based on LBP achieved 95.9\% accuracy, and \cite{4631331}, that achieved a 98.2\% accuracy by using EANT2, a neuroevolution method to develop artificial neural networks for classification purposes.

Contributions of this work include the evaluation of different state-of-the-art CNNs and the proposal of two novel methods of performing surface defect detection. Moreover, the major contribution of this work, Auto-Classifier, proposes a new way of classifying images by improving the CNNs classification mechanism by using automated search to generate a new classifier that is then coupled to the CNN. To the best of our knowledge, the Auto-Classifier method proposed in this paper is the first method that combines and improves CNNs using AutoML.

\section{Proposed Method}
\label{sec:proposed}
In this section, we present our proposed methods, which aim at classifying the presence of defects in 2D images by CNNs. We evaluate the performance of multiple state-of-the-art architectures, that yield the best results for multiple problems in the task of image and objects classification, in the task of detecting defects, which are: VGG11, VGG16 and VGG19 \cite{Simonyan15}, Resnet18, Resnet34, Resnet50 and Resnet101 \cite{7780459}, and Densenet121 \cite{8099726} (Section \ref{sec:cnns}). Based on the evaluation of these CNN architectures, we propose two methods: \emph{i)} CNN-fusion, which combines all the CNN architectures and outputs final one prediction (Section \ref{sec:cnns}); \emph{ii)} Auto-Classifier, which uses as basis, the architecture that yields the best results in the performance evaluation, and then uses Automated Machine Learning (AutoML) to automatically search a new classifier (Section \ref{sec:autoML}), that is then stacked with the CNN feature extraction mechanism.


\subsection{Convolutional Neural Architectures}
\label{sec:cnns}

CNNs are one of the most popular and prominent deep learning architectures for a variety of image processing \cite{lecun2015deep,schmidhuber2015deep,goodfellow2016deep}. CNNs are a variant of Multilayer Perceptron Networks and are biologically inspired models, created to emulate how the human visual cortex processes visual information, making these networks particularly suitable for image processing. Usually, CNNs perform a series of operations, such as convolutions and pooling, and are followed by a number of fully connected layers. The idea is: CNNs start by extracting representations of the input as features maps, which gradually increase in complexity at deeper layers of the network, then, these feature maps are fed into fully connected layers that provide the output of the network (activation patterns), normally in the form of a classification map.

Provided that CNNs can extract meaningful feature maps and thus, activation patterns, the results in a multitude of tasks are usually excellent, outperforming methods based on hand-crafted features \cite{schmidhuber2015deep}. Thus, CNNs need to be trained in a set of data so that they can extract meaningful information from the input, hence, correctly solving a given problem. The most common method to train CNNs is by using gradient-descent with Back-propagation \cite{rumelhart1986learning}, where an input image is propagated forward throughout the network, and upon reaching the final layer, the loss is calculated, and retro-propagated through the network, thus adjusting the weights of the network to more accurately solve the input problem.


In this paper, the first approach sought to evaluate the performance of several state-of-the-art CNN architectures that are known to do well in a variety of classification problems, VGG11, VGG16 and VGG19 \cite{Simonyan15}, Resnet18, Resnet34, Resnet50 and Resnet101 \cite{7780459}, and Densenet121 \cite{8099726} and combine their approach. 

By evaluating a set of CNNs (Section \ref{sec:experiments}), we created a pool of networks that were specifically trained to solve the problem of defect detection, yielding excellent results individually. To harness the classification correctness of all the individual networks, we propose CNN-Fusion. This approach intends to perform a combination of all the predictions of the individually trained CNNs, into a final, unique classification. 

Denoting that all the individual CNNs were trained using a training set and validated using a unique, validation set, all networks can be categorized by their Area Under The Roc Curve (AUC) obtained in the validation set, as metric for their correctness in solving the given problem. So, the proposed CNN-Fusion works by fusing all the individual predictions into a final, weighted, prediction by making a weighted sum of each class, where each network votes using a normalized weight based on the AUC obtained in the validation set. The weights are obtained with the following expression:
\begin{equation}
    w_i =  \frac{V_i}{\sum_{j=1}^{n}V_j}, i \in {1,...,n}
\end{equation}
where $n$ is the number of CNNs, and $V$ the array that contains the AUC values (between 0 and 1) for all the CNNs.

Hence, to perform classification, using the normalized contribution of the individual for a given input, we use the following expression:
\begin{equation}
    argmax_i (P_{ij} \cdot w_j), \begin{array}{r}i \in {1,...,c}, \\j \in {1,...,n} \end{array}
\end{equation}
where $c$ is the number of classes, $n$ the number of CNNs, $P_{ij}$ represents the output classification of network $j$ for class $i$. 

The idea behind CNN-Fusion is that, by balancing the importance of each network through the process of normalization, where networks that have higher AUC scores in the validation set have higher confidence, we can perform a weighted voting that will improve upon the result of classifying the existence of defects using individual networks.

\subsection{Auto-Classifier}
\label{sec:autoML}


\begin{figure}[!tbp]
\centering
\begin{subfigure}[b]{.48\textwidth}
  \centering
  \includegraphics[width=1\columnwidth]{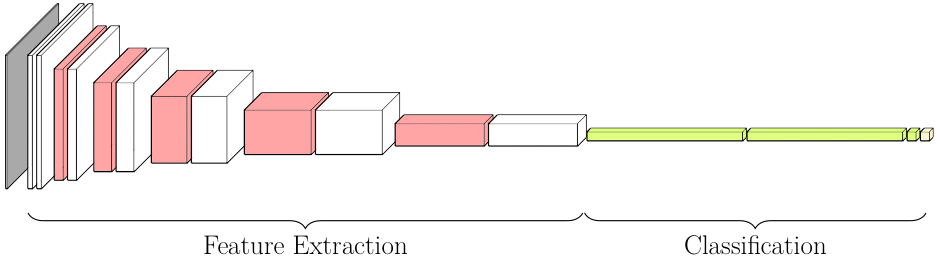}
  \caption{Individual CNN}
\end{subfigure}%
\hfill
\begin{subfigure}[b]{.48\textwidth}
  \centering
  \includegraphics[width=1\columnwidth]{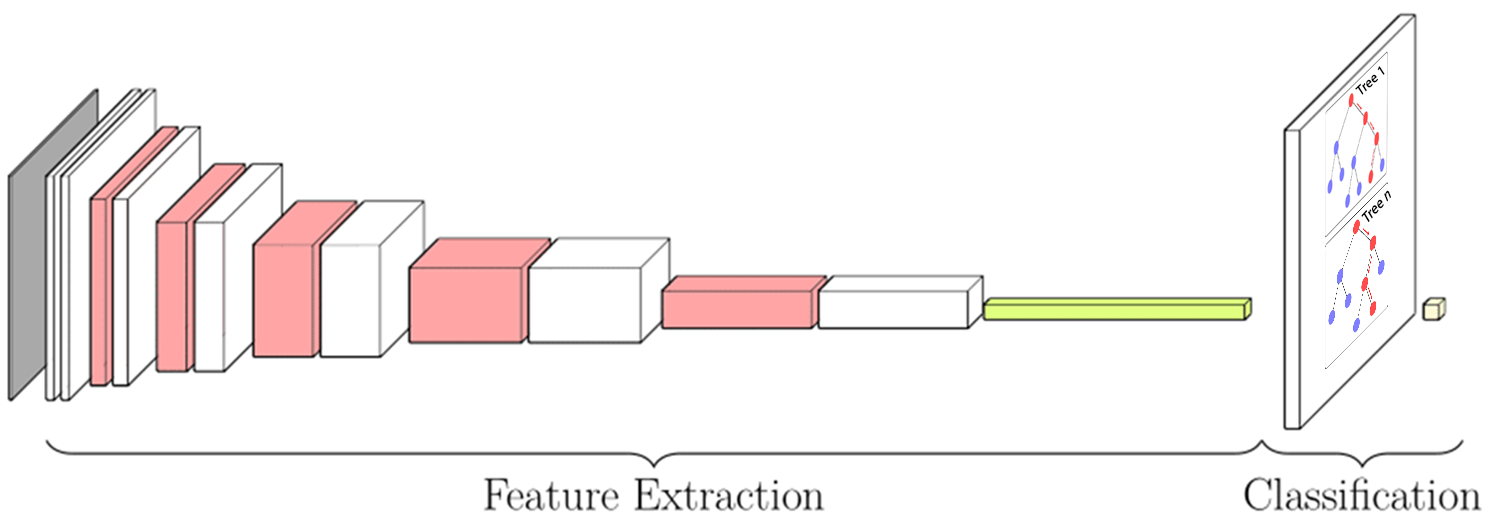}
  \caption{Auto-Classifier}
\end{subfigure}%
\caption{Visual representation of the difference between a CNN and the Auto-Classifier method. CNN is composed by two components: Feature Extraction and Classification. In the Auto-Classifier, the classification component has been replaced by another one, represented by a XGBM.\label{img:visualrepresentation}}
\end{figure}

The second proposed method in this paper, Auto-Classifier, focuses on improving the best individual CNN, by replacing its classification component by a new one. As mentioned before, CNNs are usually comprised of two parts: \emph{i)} feature extraction component, which is the initial part of the CNN, and is comprised of a set of layers, normally convolutions followed by batch-normalization, pooling layers or others, and \emph{ii)} classification component, which is the last part of a CNN and consists of a Multi-Layer Perceptron, with possible addition of regularization layers.

We hypothesize that by training a CNN from scratch and then removing its classification component partially or entirely and replacing it by other types of classification methods or even by other methods that will perform both feature extraction and classification, it might be able to outperform the initial individual CNN. This is due to the fact that other types of classifiers, such as random forests, have shown to be very good in different classification problems. The problem with many classifiers is that, to be able to process and classify images correctly, they require extensive processing power, translating into huge models. By processing the image with the feature extraction of a CNN, complex feature maps are created, which can then be used by a classifier, without need to perform any more feature extraction or feature processing. So, the proposed method, Auto-Classifier, works by initially using an individual CNN, from which the classification component is partially removed. Then, we use AutoML to generate a new classifier, based on the output features of the modified CNN. With this, we use the feature extraction capabilities of a CNN, and also allow the first layer of the CNN's classification component to stay, ensuring that the input to the new classifier has a smaller dimensionality, which ultimately enables more types of classifiers to work with that data. 

The Auto-Classifier is composed of two parts: 1) the best individual CNN without the classification component, leaving a trained CNN that outputs a representation map of the input; 2) conduct an automated search, for a new method to perform classification based on the representations generated in the previous step. Then, the final model is composed by the partial CNN, sequentially followed by the new classifier found by AutoML. In Fig \ref{img:visualrepresentation}, we visually present the difference between a CNN and the Auto-Classifier method. In this, the Auto-Classifier method, based on the CNN presented in (a), has a new classification component, represented by a XGBM.


To perform the AutoML search, we used H2O AutoML \cite{H2OAutoML}, which is intended to automate the machine learning workflow, by initially performing a Random Search of different models, and then performing a post-processing step by stacking the best solutions found \cite{gijsbers2019open}. The models in which the AutoML performs a hyperparameter search and tuning are: 3 pre-specified XGBoost Gradient Boosting Machine (XGBM) models, a fixed grid of Generalized Linear Models, a default Random Forest, five pre-specified H2O GBMs, a near-default Deep Neural Net, an Extremely Randomized Forest, a random grid of XGBoost GBMs, a random grid of H2O GBMs, and a random grid of Deep Neural Networks.

\section{Experiments}
\label{sec:experiments}


To evaluate our proposed methods, the performance of different CNNs, and compare them with baselines and competitive approaches, we conduct experiments on a popular set of datasets that contain surface defects, the DAGM2007 datasets. All the experiments were conducted using a computer with an NVidia GeForceGTX 1080 Ti, 16Gb of ram, 500GB SSD disk and an AMD Ryzen 7 2700 processor.

\subsection{DAGM2007}
The DAGM2007 consists of 6 different datasets, each with 1150 images, from which, 1000 images are of background textures without defects, and 150 images of one labelled defect each on the background texture. On each of these problems, we performed a stratified split into 3 sets: 70\% for the train set, 15\% for the validation set and 15\% for the test set. The train and validation set were used to train the algorithms, and the test set to evaluate the final performance of the methods. The test set was never used for training purposes, and for the proposed methods, the best individual CNN was selected based on its validation AUC. 

\subsection{Results and Discussion}
To validate the proposed methods, we conducted 3 experiments: \emph{1)} evaluate the performance of multiple state-of-the-art CNN architectures, \emph{2)} evaluate the performance of the CNN-fusion method, by fusing all individual CNNs, and \emph{3)} evaluate the performance of the Auto-Classifier method.

To evaluate the performance of state-of-the-art CNNs, we used the same settings for each one: Stochastic Gradient Descent algorithm, batch size of $10$, learning rate of $1e-3$, momentum of $0.9$, and $100$ epochs of training. To adjust the network's weights, we used back-propagation with gradient descent and Cross-Entropy loss. Subsequently to the training step, the model used for testing purposes is the one that yields the highest validation AUC while training. Denote that this testing step is only used to compare the different individual CNNs under the same conditions and is not used in any situation nor to select the best CNNs to be used in the proposed methods. The results, shown in the first 9 columns of Table \ref{tab:DAGM2007}, determined that amongst all the individual CNNs tested, VGG16 was the one with the best results - 99.9\% mean accuracy and 99.7\% mean AUC. The use of AUC as a metric for evaluating performance is extremely important because the accuracy metric is not, by its own, a good representative of a good classifier, since the dataset is not balanced and the accuracy shown does not consider that, whilst AUC-ROC is sensitive to class imbalance. 

\begin{table}[!t]
\caption{Results of different state-of-the-art CNNs architectures and the two proposed methods in the task of defect detection, using the DAGM2007 dataset with test splits. Accuracy and AUC values are shown in percentages.\label{tab:DAGM2007}}
\resizebox{\textwidth}{!}{%
\begin{tabular}{ccccccccccccccccccccc} 
\toprule
\multirow{2}{*}{Problem} & \multicolumn{2}{c}{VGG11}      & \multicolumn{2}{c}{VGG16} & \multicolumn{2}{c}{VGG19}     & \multicolumn{2}{c}{Resnet18}   & \multicolumn{2}{c}{Resnet34}   & \multicolumn{2}{c}{Resnet50}   & \multicolumn{2}{c}{Resnet101}  & \multicolumn{2}{c}{Densenet121}  &  \multicolumn{2}{c}{\textbf{CNN-Fusion}}  & \multicolumn{2}{c}{\textbf{Auto-Classifier}} \\ 
\cmidrule(lr){2-3} \cmidrule(lr){4-5} \cmidrule(lr){6-7} \cmidrule(lr){8-9} \cmidrule(lr){10-11} \cmidrule(lr){12-13} \cmidrule(lr){14-15} \cmidrule(lr){16-17} \cmidrule(lr){18-19} \cmidrule(lr){20-21}
     & Acc. & AUC & Acc. & AUC & Acc. & AUC & Acc. & AUC & Acc. & AUC & Acc. & AUC & Acc. & AUC & Acc. & AUC & Acc. & AUC & Acc. & AUC \\ \midrule
1    & 100& 100                   & 100& 100       & 85.0 & 50.0           & 100& 100                   & 100& 100                   & 100& 100                   & 100& 100                   & 100& 100  & 100 & 100     &100 & 100            \\
2    & 85.0& 50.00& 100 & 100        & 100 & 100           & 100& 100                   & 100& 100                   & 100& 100                   & 99.4& 98.1                   & 100& 100      & 100 & 100     &100 & 100            \\
3    & 100& 100                   & 100& 100    & 100 & 100              & 86.1& 53.9                   & 97.1& 90.4                   & 99.4& 98.1                   & 100& 100                   & 99.4& 98.1     & 100 & 100     &100 & 100         \\
4    & 100& 100                   & 99.4& 98.1        & 99.4 & 98.1              & 98.8& 97.7                   & 100& 100                   & 99.4& 98.1                   & 99.4& 98.1                   & 100& 100      & 100 & 100     &100 & 100        \\
5    & 98.8 & 96.2                   & 100 & 100        & 100 & 100           & 98.8 & 96.2                   & 98.8 & 96.2                   & 99.4 & 98.1                   & 99.4 & 98.1                   & 99.4 & 98.1      & 99.4 & 98.1     &100 & 100         \\
6     & 100 & 100                   & 100 & 100        & 100 & 100           & 100 & 100                   & 100 & 100                   & 98.8 & 96.2                   & 99.4 & 100                   & 98.3 & 94.2        & 100 & 100     &100 & 100     \\ \midrule
{$\mu$}    & 97.3 & 91.0                   & \textbf{99.9}           & \textbf{99.7}          & 97.4 & 91.4 & 97.3 & 91.3                   & 99.3 & 97.8                   & 99.5 & 98.4                   & 99.6 & 99.1                   & 99.5 & 98.4     & \textbf{99.9} & \textbf{99.7}      & \textbf{100} & \textbf{100}          \\ \bottomrule

\end{tabular}
}
\end{table}

We believe that the reason behind VGG16 having better results in almost all the six datasets and the best overall mean performances is due to the fact that, even if Resnet networks and Densenet are more powerful, their larger number of layers is a drawback when using small datasets, which is our case, since we are dealing with only 1150 images per dataset. Even though residual connects and short circuits in the mentioned networks can mitigate problems such as the vanishing gradient, their bigger complexity is a factor that will undermine their performance in problems were datasets are small and costly to acquire, e.g., defects in car painting.



By having the validation AUC values for each individual, we can complete the process of CNN-Fusion by combining the individual classifications into a final one, by first normalizing the validation AUC values, and then performing a voting using the CNN's classifications. The results for the CNN-Fusion are shown in 10th column of Table \ref{tab:DAGM2007}, where the mean accuracy was 99.9\% and mean AUC was 99.7\%. CNN-Fusion achieved the same mean values as VGG16. However, the difference is that CNN-Fusion was capable of perfectly identify defects in problems 1 to 4 and 6, and had miss-classifications in problem 5, while VGG16 had some miss-classifications in problem 4 and perfectly solved all the other problems. Moreover, CNN-Fusion was capable of achieving an overall high performance, but in problem 5, as many individual CNNs had classification errors, the CNN-Fusion was not capable of having 100\% AUC. This can be justified because even though normalizing AUC values across all CNNs is a mean to balance individual contributions, all the CNNs had very good performances, meaning that the normalized values will not have large differences. A possible improvement would be to perform a non-linear normalization, where better models have a more normalized difference from its neighbours when compared to un-normalized values. The problem with the use of CNN-Fusion in a fast-paced environment, e.g., quality control in industrial production lines, is that it takes more time to have a final classification, as it requires all CNNs to perform their classification, which if done in parallel, will be the maximum time, $t$, from the set of times, $T$, that contain the time taken for each CNN to perform a classification for a given input, plus the time taken, $t_{fusion}$, to perform the final classification using all individual classifications: $f_{time} = max\{T(x): x=1,..,n\} + t_{fusion}$, where $n$ represents the number of individual CNNs. The problem is that in the vast majority of the systems, conducting a forward pass in all the individual CNNs in a parallel manner is impossible. When done sequentially, the $f_{time}$ will be increasingly higher: $f_{time} = (\sum_{x=1}^{n}{T(x)}) + t_{fusion}$.

The Auto-Classifier solves the problem of having an inference time that is dependant on all individual CNNs, by using the feature extraction capabilities of only the overall best CNN on the validation set (VGG16 in our experiments) and improving its classification component. We partially removed the classification component of VGG16, by removing the last two fully connected layers, leaving only one, which was used to reduce the feature maps dimensionality to speed up the search for a new classifier and also to allow a larger pool of classifier candidates. We run AutoML for 2 hours for each problem, and at the end, we selected the best candidate on the validation set to be coupled to the modified VGG16 to create a final model - Auto-Classifier. The best classifiers from the AutoML step were: for problems 2 to 6, the classifier was a XGB model, and for problem 1 was a GBM. The results of Auto-Classifier on the test sets are shown in the last column of Table \ref{tab:DAGM2007}, and it is possible to see that it not only improved upon the individual CNN, VGG16, but it also correctly classified all data points in each one of the 6 datasets from DAGM2007.

An important aspect relevant for industrial systems is the time required to train the models, allowing quick changes, and the inference time, as real-time inference is of utmost importance. The overall mean time and standard deviation to train each CNN, was: $50.7 \pm 0.18$ minutes for VGG11, $95.2 \pm 0.33$ minutes for VGG16, $115.6 \pm 0.71$ minutes for VGG19, $17.0 \pm 0.02$ minutes for Resnet18, $28.8 \pm 0.02$ minutes for Resnet34, $44.8 \pm 0.04$ minutes for Resnet50, $71.1 \pm 0.09$ minutes for Resnet101, and $47.63 \pm 0.02$ minutes for Densenet121. From this, we can infer that the time required to train any of the CNNs is feasible in an environment with rapid changes, since the CNN that took more time to train in the experimented datasets, was VGG19, requiring less than 2 hours. Regarding CNN-Fusion, it required no further training, as it is the combination of training all individual CNNs, and for the Auto-Classifier, the AutoML component was allowed to search for a time limited to $2$ hours. By adding it to the time required by VGG16 to train, it required $235.6$ minutes in average to create the complete model. For inference times, all CNNs were capable of running in real-time, with the fastest one being Resnet18 with an inference time of $0.057\pm0.016$ seconds, and the slowest one being VGG19, with an inference time of $0.269\pm0.002$ seconds. As for VGG16, the one selected to be the feature extraction component of Auto-Classifier, it had an inference time of $0.225\pm0.002$, which is enough for detecting surface defects in real-time. CNN-fusion, in a serial manner, has an $f_{time}$ of $1.152$ seconds, which is not suitable for real-time. As for the Auto-Classifier, which consists of the partial VGG16 and the new classification component, we found that it is extremely fast, requiring only $0.001\pm0.004$ seconds to classify an image, which is justified by having an efficient new classifier that uses less, and faster operations than the removed layers.

We also compare our two proposed methods and the best individual CNN with other methods that achieve the best results in the DAGM2007 defect classification in Table \ref{tab:my-table}. In this table, it is possible to see that each one of the 3 methods studied here had the highest average accuracy, which is calculated by summing the true positive rate and true negative rate means, and divide it by two: $(TPR+TNR)/2$. It is also worth noting that our proposed method, Auto-Classifier, not only achieved a perfect classification on all DAGM2007 problems, but outperformed all other methods in this set of datasets. 

\begin{table}[t]
\centering
\caption{Comparison of different methods in the task of defect detection in DAGM2007 problems, using as metics the True Positive Rate (TPR), True Negative Rate (TNR), and Average Accuracy. Table adapted from \cite{weimer2016design}.\label{tab:my-table}} %
\resizebox{\textwidth}{!}{%
\begin{tabular}{lccccccc}
\toprule
Problem & \begin{tabular}[c]{@{}c@{}}\textbf{Auto-Classifier}\\\textbf{ (Ours)}\end{tabular} & \begin{tabular}[c]{@{}c@{}}\textbf{CNN-Fusion}\\\textbf{ (Ours)}\end{tabular} & \begin{tabular}[c]{@{}c@{}}\textbf{VGG16} \\ \textbf{(Ours)}\end{tabular} & \begin{tabular}[c]{@{}c@{}}Deep CNN\\ \cite{weimer2016design}\end{tabular} & \begin{tabular}[c]{@{}c@{}}Statistical\\ features \cite{scholz2012automated}\end{tabular} & \begin{tabular}[c]{@{}c@{}}SIFT and\\ ANN \cite{4631331}\end{tabular} & \begin{tabular}[c]{@{}c@{}}Weibull\\ \cite{timm2011non}\end{tabular} \\ \midrule
\textit{TPR (\%)} & &  &  &  &  &  &  \\
1 & 100 & 100 & 100 & 100 & 99.4 & 98.9 & 87.0 \\
2 & 100 & 100 & 100 & 100 & 94.3 & 95.7 & - \\
3 & 100 & 100 & 100 & 95.5 & 99.5 & 98.5 & 99.8 \\
4 & 100 & 100 & 99.3 & 100 & 92.5 & - & - \\
5 & 100 & 99.3 & 100 & 98.8 & 96.9 & 98.2 & 97.2 \\
6 & 100 & 100 & 100 & 100 & 100 & 99.8 & 94.9 \\[-2ex] \\
\textit{TNR (\%)} & &  &  &  &  &  &  \\
1 & 100 & 100 & 100 & 100 & 99.7 & 100 & 98.0 \\
2 & 100 & 100 & 100 & 97.3 & 80.0 & 91.3 & - \\
3 & 100 & 100 & 100 & 100 & 100 & 100 & 100 \\
4 & 100 & 100 & 100 & 98.7 & 96.1 & - & - \\
5 & 100 & 100 & 100 & 100 & 96.1 & 100 & 100 \\
6 & 100 & 100 & 100 & 99.5 & 96.1 & 100 & 100 \\[-2ex] \\
\multicolumn{2}{l}{Average Accuracy (\%)} & &  &  &  &  &  \\  
 &  \textbf{100.0} & 99.9 & 99.9 & 99.2 & 95.9 & 98.2 & 97.1 \\ \bottomrule
 
\end{tabular}
}
\end{table}

\section{Conclusions}
This paper studies how different CNNs perform in the task of detecting surface defects and proposes two methods to solve the problem: 1) CNN-Fusion, which fuses the different CNN classifications into a final one, and 2) Auto-Classifier, which is a novel method that leverages the feature extraction power of a state-of-the-art CNN and complements it by performing an automated search for a new classifier component. 

We initially conduct a study of how CNNs perform in a task that is usually solved by extracting hand-crafted features and then applying classifiers such as SVMs. This study showed how CNNs perform in detecting defects with low amounts of data points. Moreover, we propose a novel method, Auto-Classifier, that is capable of improving the performance of CNNs, and outperforming the current state-of-the-art in the task of detecting surface defects in DAGM2007 set of problems. With this, we can conclude that even though CNNs have exceptional results in a variety of image problems, they can be improved by partially replacing its classification component by other types of classifiers. Also, our experiments show that deep learning approaches for detecting surface defects, outperform traditional ones, and require no hand-crafted feature extraction, which removes problems that arise from environmental changes.

In short, the results of Auto-Classifier not only improve the state-of-the-art performance of surface defect detection in all problems of DAGM2007, but also show us that CNNs can be improved by replacing its inner classification component. As future work, the mechanism behind Auto-Classifier can be extended to new problems, and also study different changes in the classification component of CNNs to achieve the best possible performances in different problems.


%
%
%
%

\bibliographystyle{splncs04} 
\bibliography{references.bib} 

\begin{thebibliography}{10}
\providecommand{\url}[1]{\texttt{#1}}
\providecommand{\urlprefix}{URL }
\providecommand{\doi}[1]{https://doi.org/#1}

\bibitem{DBLP:conf/iclr/BakerGNR17}
Baker, B., Gupta, O., Naik, N., Raskar, R.: {Designing Neural Network
  Architectures using Reinforcement Learning}. In: {ICLR} 2017 (2017)

\bibitem{cheng2016learning}
Cheng, G., Zhou, P., Han, J.: Learning rotation-invariant convolutional neural
  networks for object detection in vhr optical remote sensing images. IEEE
  Transactions on Geoscience and Remote Sensing  \textbf{54}(12),  7405--7415
  (2016)

\bibitem{feurer2019auto}
Feurer, M., Klein, A., Eggensperger, K., Springenberg, J.T., Blum, M., Hutter,
  F.: Auto-sklearn: efficient and robust automated machine learning. In:
  Automated Machine Learning, pp. 113--134. Springer (2019)

\bibitem{garcia2018multi}
Garcia~Plaza, E., Lopez, P., Gonzalez, E.: Multi-sensor data fusion for
  real-time surface quality control in automated machining systems. Sensors
  \textbf{18}(12) (2018)

\bibitem{gijsbers2019open}
Gijsbers, P., LeDell, E., Thomas, J., Poirier, S., Bischl, B., Vanschoren, J.:
  An open source automl benchmark. In: ICMLW on Automated Machine Learning
  (2019)

\bibitem{goodfellow2016deep}
Goodfellow, I., Bengio, Y., Courville, A.: Deep learning. MIT press (2016)

\bibitem{H2OAutoML}
H2O.ai: H2O AutoML (June 2017),
  \url{http://docs.h2o.ai/h2o/latest-stable/h2o-docs/automl.html}, h2O version
  3.30.0.1

\bibitem{7780459}
{He}, K., {Zhang}, X., {Ren}, S., {Sun}, J.: Deep residual learning for image
  recognition. In: CVPR (2016)

\bibitem{8099726}
{Huang}, G., {Liu}, Z., {Van Der Maaten}, L., {Weinberger}, K.Q.: Densely
  connected convolutional networks. In: CVPR. pp. 2261--2269 (2017)

\bibitem{hutter2019automated}
Hutter, F., Kotthoff, L., Vanschoren, J.: Automated Machine Learning. Springer
  (2019)

\bibitem{kocbek2019automated}
Kocbek, S., Gabrys, B.: Automated machine learning techniques in prognostics of
  railway track defects. In: ICDMW. IEEE (2019)

\bibitem{kotthoff2017auto}
Kotthoff, L., Thornton, C., Hoos, H.H., Hutter, F., Leyton-Brown, K.: Auto-weka
  2.0: Automatic model selection and hyperparameter optimization in weka. The
  Journal of Machine Learning Research  \textbf{18}(1),  826--830 (2017)

\bibitem{lecun2015deep}
LeCun, Y., Bengio, Y., Hinton, G.: Deep learning. nature  \textbf{521}(7553)
  (2015)

\bibitem{li2017recognition}
Li, G., Zhao, X., Du, K., Ru, F., Zhang, Y.: Recognition and evaluation of
  bridge cracks with modified active contour model and greedy search-based
  support vector machine. Automation in Construction  \textbf{78},  51--61
  (2017)

\bibitem{liu2018darts}
Liu, H., Simonyan, K., Yang, Y.: {{{DARTS}: Differentiable Architecture
  Search}}. In: ICLR (2019)

\bibitem{malamas2003survey}
Malamas, E.N., Petrakis, E.G., Zervakis, M., Petit, L., Legat, J.D.: A survey
  on industrial vision systems, applications and tools. Image and vision
  computing  (2003)

\bibitem{mendoza-automlbook18a}
Mendoza, H., Klein, A., Feurer, M., Springenberg, J.T., Urban, M., Burkart, M.,
  Dippel, M., Lindauer, M., Hutter, F.: Towards automatically-tuned deep neural
  networks. In: AutoML: Methods, Sytems, Challenges (Dec 2018)

\bibitem{mital1998comparison}
Mital, A., Govindaraju, M., Subramani, B.: A comparison between manual and
  hybrid methods in parts inspection. Integrated Manufacturing Systems  (1998)

\bibitem{ENAS}
Pham, H., Guan, M., Zoph, B., Le, Q., Dean, J.: {{Efficient Neural Architecture
  Search via Parameters Sharing}}. In: ICML (2018)

\bibitem{rumelhart1986learning}
Rumelhart, D.E., Hinton, G.E., Williams, R.J.: Learning representations by
  back-propagating errors. nature  \textbf{323}(6088),  533--536 (1986)

\bibitem{schmidhuber2015deep}
Schmidhuber, J.: Deep learning in neural networks: An overview. Neural networks
   \textbf{61},  85--117 (2015)

\bibitem{7926704}
{Schmugge}, S.J., {Rice}, L., {Lindberg}, J., {Grizziy}, R., {Joffey}, C.,
  {Shin}, M.C.: Crack segmentation by leveraging multiple frames of varying
  illumination. In: WACV (March 2017)

\bibitem{scholz2012automated}
Scholz-Reiter, B., Weimer, D., Thamer, H.: Automated surface inspection of
  cold-formed micro-parts. CIRP annals  \textbf{61}(1),  531--534 (2012)

\bibitem{4631331}
{Siebel}, N.T., {Sommer}, G.: Learning defect classifiers for visual inspection
  images by neuro-evolution using weakly labelled training data. In: IEEE CEC
  (2008)

\bibitem{Simonyan15}
Simonyan, K., Zisserman, A.: Very deep convolutional networks for large-scale
  image recognition. In: ICLR (2015)

\bibitem{sohn2012learning}
Sohn, K., Lee, H.: Learning invariant representations with local
  transformations. In: ICML (2012)

\bibitem{timm2011non}
Timm, F., Barth, E.: Non-parametric texture defect detection using weibull
  features. In: Image Processing: Machine Vision Applications IV (2011)

\bibitem{villalba2019deep}
Villalba-Diez, J., Schmidt, D., Gevers, R., Ordieres-Mer{\'e}, J., Buchwitz,
  M., Wellbrock, W.: Deep learning for industrial computer vision quality
  control in the printing industry 4.0. Sensors  \textbf{19}(18), ~3987 (2019)

\bibitem{weimer2016design}
Weimer, D., Scholz-Reiter, B., Shpitalni, M.: Design of deep convolutional
  neural network architectures for automated feature extraction in industrial
  inspection. CIRP Annals  \textbf{65}(1),  417--420 (2016)

\bibitem{wieler2007weakly}
Wieler, M., Hahn, T.: Weakly supervised learning for industrial optical
  inspection. In: DAGM symposium in 2007 (2007)

\bibitem{xie2008review}
Xie, X.: A review of recent advances in surface defect detection using texture
  analysis techniques. ELCVIA  \textbf{7}(3),  1--22 (2008)

\bibitem{zela2020understanding}
Zela, A., Elsken, T., Saikia, T., Marrakchi, Y., Brox, T., Hutter, F.:
  {Understanding and Robustifying Differentiable Architecture Search}. In: ICLR
  (2020)

\bibitem{zhang2016road}
Zhang, L., Yang, F., Zhang, Y.D., Zhu, Y.J.: Road crack detection using deep
  convolutional neural network. In: ICIP (2016)

\bibitem{DBLP:conf/cvpr/ZhongYWSL18}
Zhong, Z., Yan, J., Wu, W., Shao, J., Liu, C.L.: {Practical Block-Wise Neural
  Network Architecture Generation}. In: CVPR (2018)

\bibitem{DBLP:journals/corr/ZophL16}
Zoph, B., Le, Q.V.: Neural architecture search with reinforcement learning. In:
  ICLR (2017)

\bibitem{Zoph_2018}
Zoph, B., Vasudevan, V., Shlens, J., Le, Q.V.: Learning transferable
  architectures for scalable image recognition. CVPR  (2018)

\end{thebibliography}

\end{document}